\documentclass[conference]{IEEEtran}

\usepackage{balance}

\IEEEoverridecommandlockouts 
\usepackage[pdftex]{graphicx}
\usepackage{algorithmic}
\usepackage{algorithm}
\usepackage{listings}
\usepackage[OT4,T1]{fontenc}
\usepackage[cmex10]{amsmath}
\usepackage{amssymb}
\usepackage{url}
\usepackage{multirow}
\usepackage{color}
\usepackage[caption=false]{subfig}

\title{Superiority of Simplicity: A Lightweight Model for Network Device Workload Prediction}
\author{\IEEEauthorblockN{
  Alexander Acker\IEEEauthorrefmark{1}\IEEEauthorrefmark{2},
  Thorsten Wittkopp\IEEEauthorrefmark{1}\IEEEauthorrefmark{2},
  Sasho Nedelkoski\IEEEauthorrefmark{1},
  Jasmin Bogatinovski\IEEEauthorrefmark{1},
  Odej Kao\IEEEauthorrefmark{1}}
  \IEEEauthorblockA{
    \IEEEauthorrefmark{1}
    Technische Universit\"at Berlin, Germany\\
    \{alexander.acker, t.wittkopp, sasho.nedelkoski, jasmin.bogatinovski, odej.kao\}@tu-berlin.de
  }
  \IEEEauthorblockA{
    \IEEEauthorrefmark{2}
    Alphabetic order, equal contribution
    }
}

\begin{document}
\maketitle              

\begin{abstract}
The rapid growth and distribution of IT systems increases their complexity and aggravates operation and maintenance. 
To sustain control over large sets of hosts and the connecting networks, monitoring solutions are employed and constantly enhanced. 
They collect diverse key performance indicators (KPIs) (e.g. CPU utilization, allocated memory, etc.) and provide detailed information about the system state.
Storing such metrics over a period of time naturally raises the motivation of predicting future KPI progress based on past observations.
This allows different ahead of time optimizations like anomaly detection or predictive maintenance.
Predicting the future progress of KPIs can be defined as a time series forecasting problem.
Although, a variety of time series forecasting methods exist, forecasting the progress of IT system KPIs is very hard.
First, KPI types like CPU utilization or allocated memory are very different and hard to be modelled by the same model.
Second, system components are interconnected and constantly changing due to soft- or firmware updates and hardware modernization.
Thus a frequent model retraining or fine-tuning must be expected.
Therefore, we propose a lightweight solution for KPI series prediction based on historic observations.
It consists of a weighted heterogeneous ensemble method composed of two models - a neural network and a mean predictor.
As ensemble method a weighted summation is used, whereby a heuristic is employed to set the weights.
The lightweight nature allows to train models individually on each KPI series and makes model retraining feasible when system changes occur.
The modelling approach is evaluated on the available FedCSIS 2020 challenge dataset and achieves an overall $\textbf{R}^{\textbf{2}}$ score of $\textbf{0.10}$ on the preliminary $\textbf{10\%}$ test data and $\textbf{0.15}$ on the complete test data.
We publish our code on the following github repository: \textit{https://github.com/citlab/fed\_challenge}

\end{abstract}

\section{Introduction}\label{intro}
\IEEEoverridecommandlockouts\IEEEPARstart{I}{T} systems are rapidly evolving to meet the growing demand for new applications and services in a variety of fields like industry, medicine or autonomous transportation. 
This entails an increasing number of interconnected devices, large networks and growing data centres to provide the required infrastructure. 
Although accelerating innovations and business opportunities, this trend increases complexity and thus, aggravates the operation and maintenance of these systems.
Operators are in need of assistance to be able to maintain control over this complexity.
Therefore, monitoring solutions are implemented.
They constantly collect system KPIs like latency, throughput, or system resource utilization and provide detailed information about the monitored IT system.
One particularly important aspect of system monitoring is the prediction of future system load based on historic observations.
Several efforts where made to enable this ranging from linear regression~\cite{dinda2000host}, Bayesian statistics~\cite{di2012host} and neural networks~\cite{song2018host}.

A precise prediction of future system load enables predictive decision making and thus, ahead of time optimization.
An anomaly detection methods can be employed to compare the difference between the predicted and the actual state and raise alarms in case of unforeseen deviations~\cite{schmidt2018unsupervised}.
Scaling up on imminent load peaks or scaling down during moderate or low utilization periods helps to optimize for cost and optimal user experience~\cite{mao2010cloud}.
Among many others, scheduling decision~\cite{jiang2012joint}, network routing and dimensioning~\cite{howard2018inverted}, data centre cooling control~\cite{ahmad2010joint} or predictive maintenance~\cite{yaseen2017iot} all benefit from precise system load predictions.

A particular data source for load prediction are KPIs like CPU utilization, allocated memory or network throughput of individual system components. 
Sampled at fixed time intervals or aggregated over a period, they represent the evolution of system states as time series.
Following this, the task of system load prediction can be formulated as a time series forecasting problem.
Forecasting different types of KPI time series based on historic observations is very hard due to the properties of IT systems.
First, different KPI types are highly non-uniform.
While CPU utilization is usually very volatile, memory allocation is rarely overlaid by noise.
Further, disk read and write operations expose bursty patterns due to buffering resulting in flat sequences with sporadic peaks.
The concrete pattern of these series depend of unknown external and a variety of internal factors.
An example of an external factor is the difference of user behaviours based on days of weeks, night- and daytime ours or occasional events like Christmas days.
Load on weekends can significantly differ from the load on weekdays.
The same applies to the difference between night and day time hours.
Therefore, load in general usually follows seasonal patterns and long term trends.
Incorporating this information either explicit or implicit into the prediction model enhances the prediction performance.
Also, the IT system itself is problematic from modeling perspective due to their dynamic nature and high uncertainty.
Frequent soft- and firmware updates or hardware modernization change system properties and usually require model retraining or fine-tuning.
It does not only affect the changed component itself but might also propagate to connected devices.
Co-locations, changing scheduling policies and maintenance operations are additional sources of uncertainty.
This imposes the requirement of frequent and fast model adaption.

Related work on time series forecasting is diverse and ranges from traditional linear or non-linear regression~\cite{stock1998comparison}, stochastic methods~\cite{hassan2005stock}, deep learning models~\cite{lim2019temporal} and ensemble methods~\cite{zhang2003time, qiu2017empirical}.
Traditional regressive or statistical models are often not able to capture the underlying complex processes which result in imprecise predictions.
Models based on neural networks or ensemble methods usually provide more accurate predictions but suffer from high complexity and an accompanying high computational overhead that is required to train them.
This is a major limitation when it comes to systems with large amounts of components as data centres or large networks.
Additionally, the previously described dynamic nature aggravates this limitation.

Considering this, we present our solution for this years FedCSIS 2020 challenge. 
It proposes a model for network device workload prediction whereby future KPI values have to be predicted based on historic data.
Due to the high number of components and KPI time series, we focused on a lightweight modelling approach in order to keep the solution computationally feasible.
The model combines the overall average of each time series with a prediction from a linear neural network.
Furthermore, we employed heuristics to tackle numerical imprecision and enhance overall prediction performance.
Our solution achieved an overall $R^2$ score of $0.10$ on the preliminary 10\% test data and $0.15$ on the complete test data.

The rest of the paper is structured as follows. 
First, section~\ref{sec:wlpred} described the problem of network device workload prediction and provides a preliminary analysis of the available training data set.
Second, section~\ref{sec:main} introduces our solution for workload prediction. It includes a formal problem definition based on time series forecasting and explains each element of our proposed method. Third, an evaluation of our method with respect to runtime and prediction performance is performed. The results are presented in section~\ref{sec:eval}. Finally section~\ref{sec:conclusion} concludes our paper.

\section{Network Device Workload Prediction} \label{sec:wlpred}

This year FedCSIS 2020 challenge was to predict the future workload of network devices based on past workload observations.
More specifically, the workload of a set of devices, referred to as hosts, were characterized by KPI series such as CPU utilization, incoming and outgoing network traffic or allocated main memory.
The data were collected hourly over a period of 3 months with sporadically missing samples.
Overall, 45 different KPIs were recorded from 3,716 hosts, whereby the workload of individual hosts was described by different KPI subsets.
Each hourly KPI series sample consists of seven measurement aggregations over the respective hour.
These are the number of collected measurements, the mean and standard deviation, the first, last, highest and lowest measurement.
All of the seven aggregations can be used as input but only the future mean value must be predicted, resulting in a possibly multivariate input but univariate output.

\begin{figure*}%
    \centering
    \subfloat{{\includegraphics[width=\columnwidth]{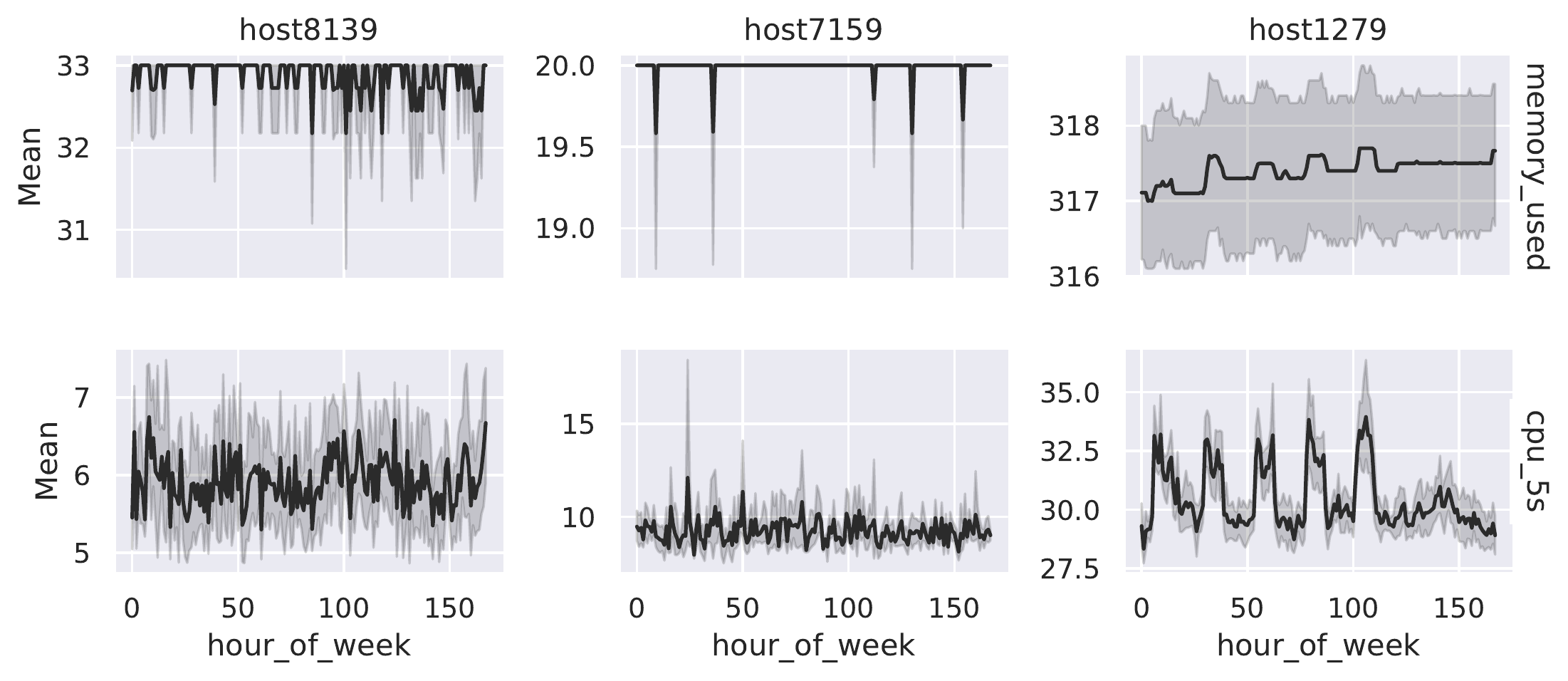}}}%
    \subfloat{{\includegraphics[width=\columnwidth]{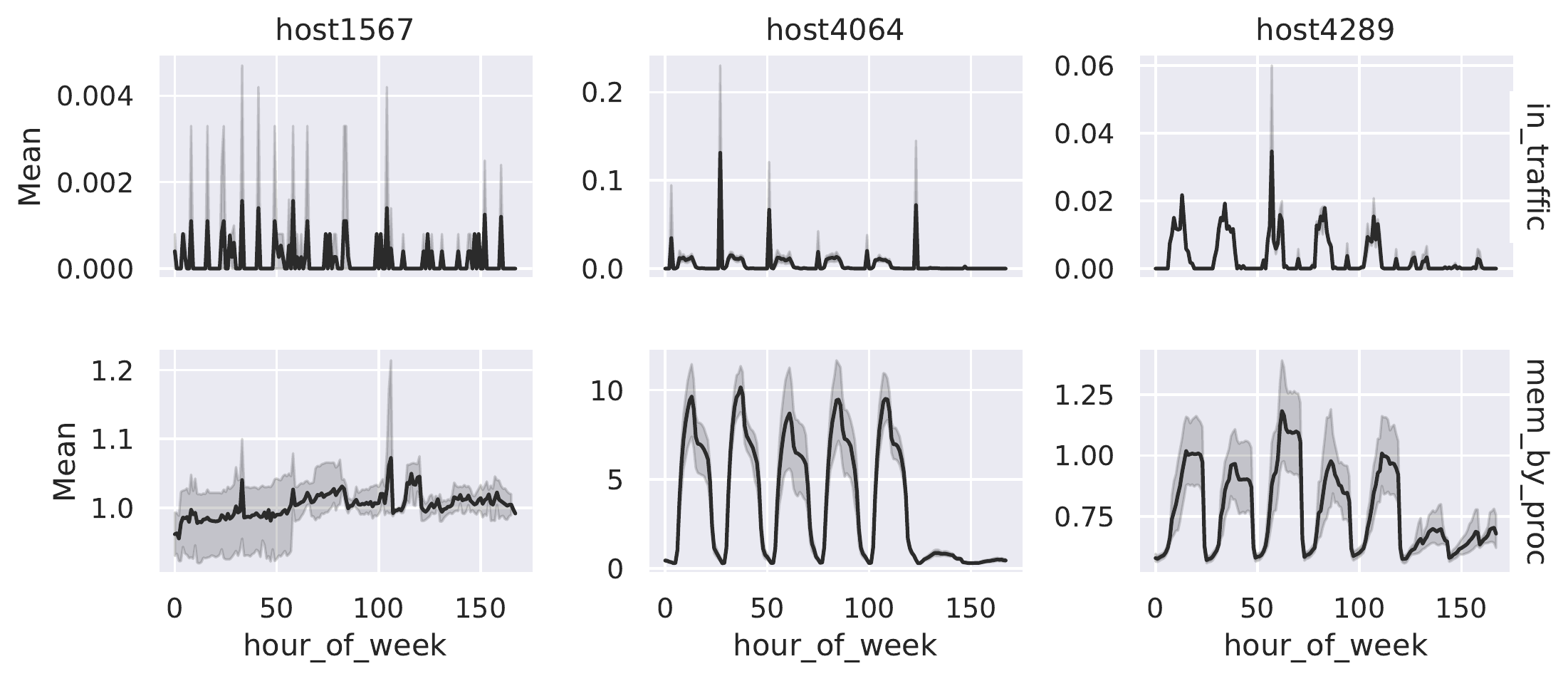}}}%
    \caption{Example of four KPIs for six hosts. A great in-between and within KPI value diversity for the different hosts can be observed. This indicates the major challenge when faced with forecasting the expected future values of the KPIs.}%
    \label{fig:KPIs}%
\end{figure*}

The plots in \figurename~\ref{fig:KPIs} show four different KPI mean values from six different hosts.
Thereby, the series was split into weekly windows from Monday until Sunday and arranged by the hour of the week resulting in ten aggregated weekly series for each plot.
The dark blue line shows the mean value while the light blue is visualized the $0.95$ confidence interval.
It can be observed that KPI series are highly non-uniform, which indicates the major challenge when faced with forecasting the expected future values of the KPIs.
There are KPI series with high noise ("cpu\_5s" of host 8139) while others remain fairly constant ("memory\_used" of host 7159).
Some KPIs follow long term trends ("memory\_used" of host 1279) and several series are periodical on a daily and weekly basis ("mem\_by\_proc" and "in\_traffic" of hosts 4064 and 4289).
Furthermore, these properties vary for the same KPI types depending on the host from which they were collected.
While "cpu\_5s" is fairly constant but noisy for host 7159, a clear seasonality can be observed for host 1279.

\section{Lightweight Workload Prediction Model} \label{sec:main}
In this section we present our method for lightweight workload prediction. Its concept and architecture were chosen based on the previously described observations and analyses in section~\ref{sec:wlpred}. In subsection~\ref{subsec:prel} we describe the workload prediction problem in form of time series forecasting and define all required preliminaries. After that, we present our method for workload prediction in subsection~\ref{subsec:model} including the data preprocessing, feature selection and forecasting.

\subsection{Preliminaries}\label{subsec:prel}

We define the task of workload prediction as a time series forecasting problem. 
A time series is an temporally ordered sequence of values $X = ({X}_t(\cdot) \in \mathbb{R}^d : t=1,2,\ldots, T)$, where $d$ is the dimensionality of each point.
For $X^{a}_{b}(\cdot)=(X_a(\cdot), X_{a+1}(\cdot), \ldots, X_b(\cdot))$, we denote indices $a$ and $b$ with $a \leq b$ and $0 \leq a,b \leq T$ as time series boundaries in order to slice a given series $X^{0}_{T}(\cdot)$ and acquire a subseries $X^{a}_{b}(\cdot)$.
The variable $T$ defines the time stamp of the last sample of the past observations.
Additionally, we use the notion $X(i)$ to refer to a certain dimension $i$, with $1 \leq i \leq d$.
Furthermore, meta information for each time series value ${X}_t(\cdot)$ are denoted as $M_t$.

The problem of workload prediction is modelled as the forecasting of a future univariate value $X_{T+w}(i)$, with $w \geq 1$, conditioned on a sequence of past values $X^{0}_{T}(\cdot)$, and known meta information about the future time stamp $M_{T+w}$.
Therefore, the learning objective is to select a function $h:\mathbb{R}^N \mapsto \mathbb{R}$, where $N$ is the dimensionality of the input, that results in a small generalization loss:
\begin{equation}
     \mathcal{L} = \frac{1}{|\mathcal{W}|}\sum_{w \in \mathcal{W}} L(h(X^{0}_{T}(\cdot), M_{T+w}), X_{T+w}(\cdot)).
\end{equation}
Thereby, $L$ is a bounded loss function and $ \mathcal{W}$ is the set of offsets defining all future time stamps to predict.

\subsection{Lightweight Workload Prediction Model} \label{subsec:model}

\begin{figure}[htbp]
\centering
\includegraphics[width=\columnwidth]{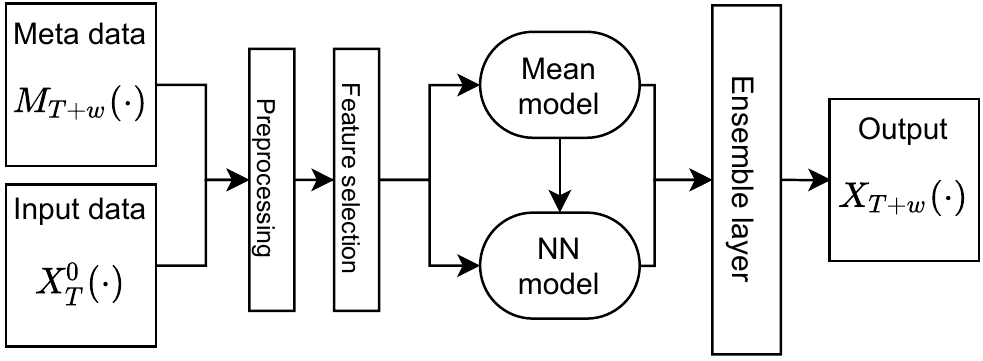}
\caption{Overall solution architecture.}
\label{model}
\end{figure}

The overall architecture of our method is depicted in \figurename~\ref{model}.
A future time series value $X_{T+w}(i)$ should be predicted based on the history $X^{0}_{T}(\cdot)$ and its known meta information $M_{T+w}$.
For the task of workload prediction, each time series $X$ represents an KPI. 
The respective dimensions of samples $X_{t}(\cdot)$ are aggregated values of that KPI between time $t-1$ and $t$.
Due to their importance, we selectively define the mean and last measurement as $\overline{x}_t$ and $x^{(l)}_t$, where $\overline{x}_{t}, x^{(l)}_t \in X_{t}(\cdot)$.
The mean value of the sample $\overline{x}_{T+w} \in X_{T+w}(\cdot)$ is the prediction target.
Since many  workload series are seasonal, we additionally add the encoded day of week and hour of day as meta information $M_{T+w}$.
Subsequently, each model element is described in detail.

\textbf{Preprocessing.} Initially, a rescaling of each value in the KPI series $X^{0}_{T}(\cdot)$ to a uniform range $[c, d]$ is performed.  
Furthermore, values in $X^{0}_{T}(\cdot)$ are expected to be sampled hourly. 
If samples are missing, a linear interpolation is employed.

\textbf{Feature Selection.} Due to the additional overhead that is introduced by automated feature selection methods, we choose to select a fixed subset of features manually.
Furthermore, we focus on a minimal set of features to keep the model capacity low.
The features are selected depending on the model that they are forwarded to.
Therefore, we define a filter $F_1$ for the mean predictor and a filter $F_2$ for the neural network model (NN).
The filter $F_1$ includes only the mean values of $X^{0}_{T}(\cdot)$.
Filter $F_2$ applies two feature selection operations.
First, out of the aggregated values in the last available series sample, we pick the mean and last value, i.e. $\overline{x}_T, x^{(l)}_T \in X_{T}(\cdot)$.
Second, motivated by the seasonality of system load, we additionally use the mean value of the same hour of the week as the prediction target of previous $k$ weeks.

\textbf{The Models.} The mean predictor calculates the overall average over the filtered sample series $F_1(X^{0}_{T}(\cdot))$.
The NN model is a linear feed-forward neural network. It receives the preprocessed and filtered data $F_2(X^{0}_{T}(\cdot))$, the meta-information values $M_{T+w}$ and the output of the mean model.
These are combined to a flat input vector $\textbf{x}$.
The learning objective of the NN model is to minimize the squared error loss between the prediction and the mean value of $\overline{x}_{T+w} \in X_{T+w}(\cdot)$:
\begin{equation}
     L = (h(\textbf{x}) - \overline{x}_{T+w})^2.
\end{equation}

\begin{figure}[htbp]
\centering
\includegraphics[width=\columnwidth]{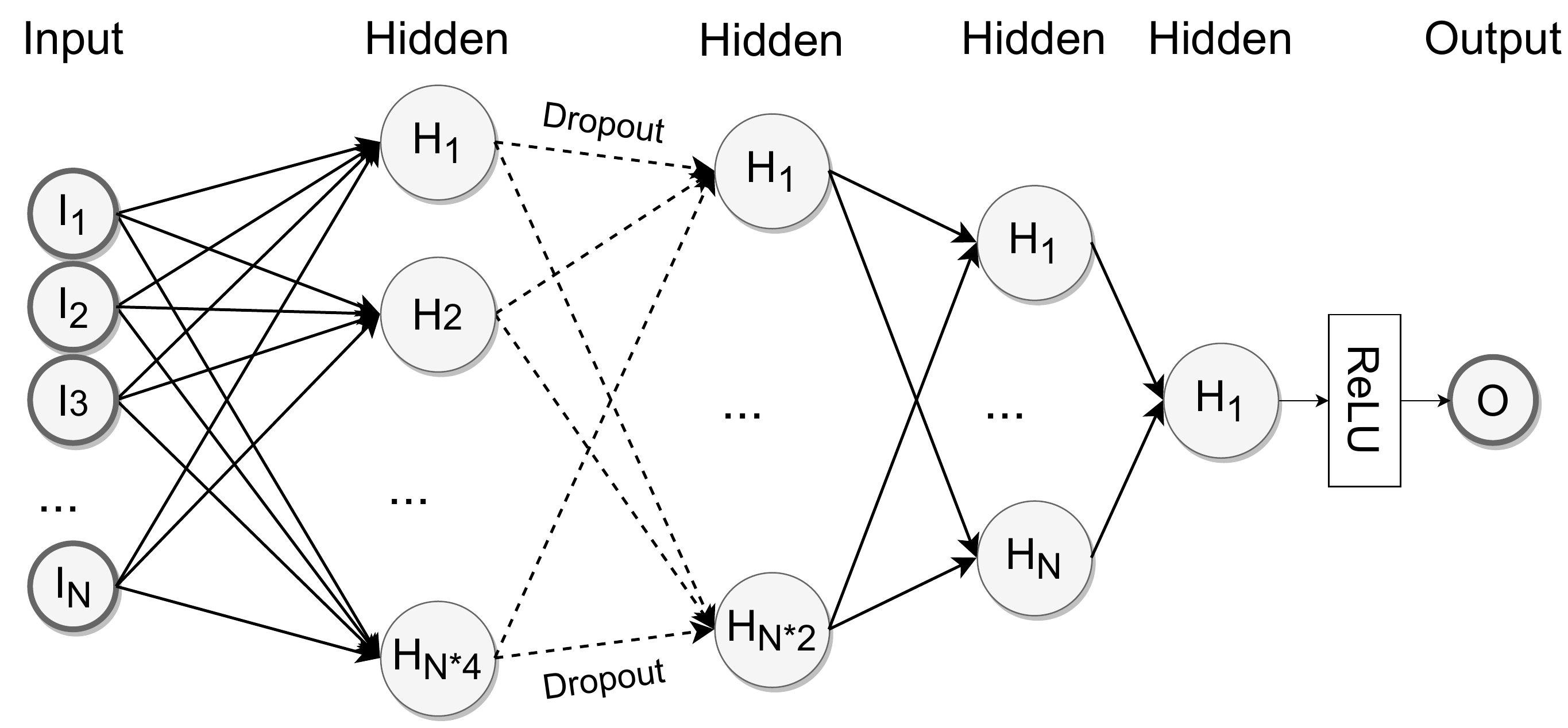}
\caption{Structure of our fast forward network.}
\label{fast_forward_net}
\end{figure}

Our employed network structure is depicted in \figurename~\ref{fast_forward_net}.
We use a fanning out first hidden layer. 
Its size is fourfold of the input layer size.
The subsequent layers are tampered, which works as regularization.
Furthermore, we use a dropout between the first and second hidden layer as an additional regularization.
A rectifier linear unit (ReLU) activation is applied to the output value of the network. 
The output of the mean model and NN model are respectively denoted as $o^{(1)}_{T+w}$ and $o^{(2)}_{T+w}$.

\textbf{Ensemble Layer.} To combine the predictions of the mean model and the NN model, a weighted average over the model outputs is calculated: 
\begin{equation}
     o_{T+w}=\sum_{o^{(i)}_{T+w} \in \{o^{(1)}_{T+w}, o^{(2)}_{T+w}\}}w_i o^{(i)}_{T+w}, \text{where} \sum_i w_i = 1.
\end{equation}
The usage of two models is motivated by the non-uniformity of KPI series.
While the neural network is capable to predict seasonal series fairly well, it fails to accurately predict constant but noisy series.
A simple average over all mean metrics of a KPI resulted in good predictions for constant but noisy series but resulted in bad predictions for seasonal series.
By combining both, we expect to achieve a generally better result.

\section{Evaluation} \label{sec:eval}
Given three months of historic data, the task is to predict respective mean KPI values of the subsequent week.
Overall the future progress of 10,000 KPI series mus be predicted.
Samples are given hourly and thus, the predictions are expected hourly as well.
This results in a sequence of 168 samples that have to be predicted for each series.
In this section, we evaluate the proposed method in terms of runtime and prediction performance.
Subsection~\ref{subsec:par_train} described the parametrization that was used to predict the submission results together with the training process.
In subsection~\ref{subsec:runtime} we provide a runtime analysis underlining the requirement of a lightweight modeling solution.
Finally, subsection~\ref{subsec:pred_res} provides an overview of the achieved challenge scores.

\subsection{Training and Parameterization} \label{subsec:par_train}
KPI series are diverse depending on the type and the host from which they were collected.
Therefore, we choose to train individual models for each KPI series.
The mean model calculates an overall average over all mean values from the available three months of data.
This, filter $F_1$ selects all available mean values $\overline{x}_t$ of the respective KPI series.

Training of the NN model requires the definition of a training set.
Therefore, a set of inputs and prediction targets are defined.
The target is always a specific mean value $\overline{x}_{t_{p}} \in X_{t_p}(\cdot)$ at prediction target time stamp $t_p \leq T$.
As target value meta information its hour of day and day of week is used, defined as $M_{t_p}=\{m_1, m_2\}$, where $m_1 \in \{1,2,\ldots,24\}$ is the hour of day and $m_2 \in \{1,2,\ldots,7\}$ is the day of week.
To acquire the input data, a filter $F_2$ is utilized on $k$ preceding weeks.
This KPI training series slice is defined as $X^{s}_{e}(\cdot)$ with $e=t_p-(m_2*24+m_1)$ and $s=e-168*k$, where $168$ are the hours of one week, $s \geq 0$ and $k \geq 1$.
Thereof, the mean and last value from the last sample are selected $\overline{x}_e, x^{l}_e \in X_{e}(\cdot)$.
Further, respecting the seasonality of several KPI series, the mean value of the same hour of the week as the prediction target is added to the input.
These can be accessed via $\{\overline{x}_{\tau} \in X_{\tau}(\cdot): \tau = t_p-i*168, i=1,2,\ldots,k\}$

To create the training data we set $k=2$.
For the rescaling, we define $c=0$ and $d=100$.
The neural network is trained via backpropagation.
Thereby, the mean square error is used as the optimization criterion and Adam as the optimizer.
We set the learning rate to $10e^-3$ and use dropout probability of $0.1$.
One individual model is trained for every KPI series.
We set a fixed number of six epochs and do not use a validation set to make use of all available data for training.
Based on the above definition of training data creation, all possible input/target tuples were used and defined as one epoch of the training process.

\subsection{Runtime Analysis} \label{subsec:runtime}

\begin{figure}[htbp]
\centering
\includegraphics[width=0.8\columnwidth]{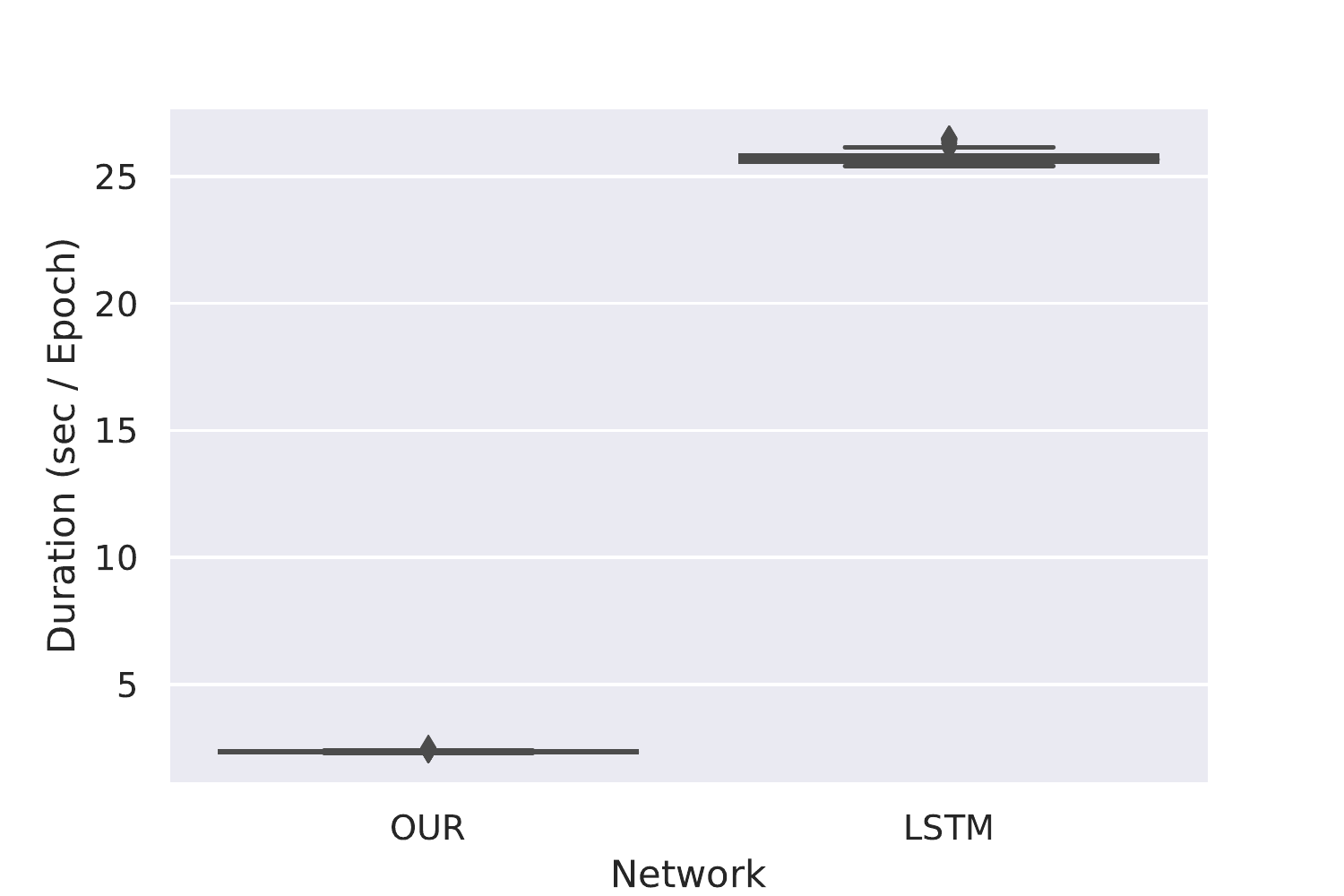}
\caption{Runtime analysis results.}
\label{fig:runtime_lstm}
\end{figure}

Due to their non-uniformity, we propose to train models respectively for each KPI series. 
Furthermore, frequent retraining can be expected due to the dynamic nature of IT systems. 
Considering this, we conduct a preliminary runtime analysis and compare our neural network to a recurrent version of it.
For the recurrent network, we use long short term memory (LSTM) instead of linear cells.
The overall architecture remains the same as depicted in \figurename~\ref{fast_forward_net}.
We measure the training time per epoch on a bare-metal machine with an Intel(R) Core(TM) i5-9600K CPU @ 3.70GHz, 3x32 GB RAM and two Nvidia GeForce RTX 2080 Titan GPUs whereof one was utilized during the runtime measurement experiments.
Ubuntu 18.04.3 LTS with kernel version 5.3.0-51-generic is installed as OS and Python version 3.6.7 and PyTorch version 1.4.0 are used to implement the networks.
The result of the comparison is shown in \figurename~\ref{fig:runtime_lstm}.

It can be observed that the LSTM version requires significantly more time for training than the network with linear cells. 
In comparison, the runtime increases by a factor of ten.
The mean training runtime per epoch of the linear version is $2.37$ seconds per epoch with a standard deviation of $0.03$ and $0.95$ confidence interval of $[2.38, 2.37]$.
For the network version with LSTM nodes a training time per epoch of $25.72$ is measured with a standard deviation of $0.18$ and $0.95$ confidence interval of $[25.74, 2.70]$.
Having six epochs per series and a total number of $10,000$ series to predict results in a total required training time of $39.5$ hours for the linear version and $17.9$ days when using LSTM cells.

Although recurrent neural network architectures especially with LSTM cells are reported to perform well on sequential data prediction tasks~\cite{nedelkoski2019anomaly}, our runtime analysis shows that the required training time is very high.
The task of training a model for each series is completely parallelizable. 
However, our access is limited to the above describe machine and the training time of almost $18$ days is infeasible for us. 
Therefore, the utilization of linear cells is chosen.

\subsection{Prediction Results} \label{subsec:pred_res}
The performance of the proposed workload prediction method is evaluated against the withheld test set by submitting the solution via the official FedCSIS 2020 challenge submission system.
The submissions are scored by the $R^2$ score defined as
\begin{equation}
     R^2 = 1 - \frac{\sum_i (\overline{x}_t - o_t)^2}{\sum_i (\overline{x}_t - \overline{\overline{x}})^2}, 
     \label{eq:r2}
\end{equation}
where $\overline{x}_t \in X_t(\cdot)$ and $\overline{\overline{x}}$ as the overall average over all mean samples.
Based on our observation several KPI series are mainly constant with sporadic deviations, resulting in a very small normalization value (denominator of in Eq.~\ref{eq:r2}).
This results in high division values and thus, low $R^2$ scores even for small deviations of the predicted values.
These values had a high impact on the overall $R^2$ score.
Furthermore, several KPI series can be described as the noise around a baseline. 
Such series are better predicted by their baseline instead of guessing random noise.
This motivates us to implement a heuristic to choose an adaptive weighting of the model outputs. 
Either an equal withing of 0.5 for each model is set, or we set the NN model weight to 0.0 and the average predictor output weight to 1.0
The decision is made as follows.
First, the neural network is trained.
Second, the last available week is used as a prediction target and the respectively filtered data before that week as input.
Since this last week was explicitly trained on, we assume precise prediction results, i.e. $R^2$ score close to 1. 
If the neural network output resulted in a lower score than the output of the average predictor, we set the weight for the average predictor to 1.0 and the neural network weight to 0.0.
Otherwise, the both weights were set to 0.5.

Finally, the prediction of the submission is done based on the filtered $k$ last available weeks in the training data set.
The $R^2$ score results are listen in \tablename~\ref{tab:results}.
\begin{table}[]
\caption{$R^2$ scores of best three submissions together with the baseline.}
\begin{tabular}{|c|c|c|c|c|}
\hline
                            & baseline & 1st    & 2nd    & Ours    \\ \hline
Preliminary test set (10\%) & 0.2267   & 0.1888 & 0.1841 & 0.1053  \\ \hline
Complete test set (100\%)   & 0.2295   & 0.163  & 0.1515 & 0.1501  \\ \hline
\end{tabular}
\label{tab:results}
\end{table}
None of the submitted results is able to achieve the specified baseline.
Two submissions achieved a better $R^2$ score than our solution with $0.1888$ and $0.1841$ on the preliminary $10\%$ of test data and $0.163$ and $0.1515$ on the complete test dataset.
With our proposed lightweight model, we achieve an $R^2$ score of $0.1053$ on the preliminary $10\%$ test data and $0.1501$ on the complete test dataset.
We did not carry out any attempts to optimize for the $10\%$ preliminary test data since it was not clear whether it is a general representation of the complete test dataset.
Therefore, it is interesting for us to see that our solution is the only one - also including submission below ours - achieving a better score on the complete dataset than on the preliminary $10\%$.

\section{Conclusion} \label{sec:conclusion}
We tackle the given challenge of network device workload prediction based on KPI data with a lightweight model that ensembles the predictions of a multi-layer linear neural network and an overall averaging predictor.
The ensemble is done by a weighted summation.
A heuristic is used to selectively set the weights for both model predictions.
The lightweight nature of the method allows training individual models for each KPI respecting the diverse natures different KPI types and host.
From a practical perspective, frequent retraining needs to be feasible which is supported by the lightweight nature of the solution as well.

To evaluate our solution we conducted two types of experiments.
First, we evaluate our solution with the FedCSIS 2020 challenge dataset.
It consists of 45 different KPIs recorded from 3,716 hosts.
The experiment results show that the lightweight approach predicts future KPI values with an overall $R^{2}$ score of $0.10$ on the preliminary $10\%$ test data and $0.15$ on the complete test data.
Second, we provide a runtime analysis between LSTM and linear network cells and show that the usage of LSTM cells increases the training time by a factor of $10$ which renders it infeasible to be used for the given problem.

For future work, we see further experimentation with different network types like convolutional neural networks or attention mechanisms as promising.
Furthermore, a different numerical encoding of the currently used meta information and the learning of the summation weights when aggregating overall average and neural network outputs are sources for potential optimization.

\bibliographystyle{IEEEtran}
\bibliography{main}

\end{document}